\documentclass{article}

\usepackage[ruled,vlined]{algorithm2e}
\usepackage{amsmath}
\usepackage{amssymb}
\usepackage{subfigure}
\usepackage{tabularx}
\usepackage{theorem}
\usepackage{url}
\usepackage{graphicx}
\usepackage{amsfonts}
\usepackage[table]{xcolor}

\title{Mean-based Heuristic Search for Real-Time Planning}

\author{Damien Pellier, Bruno Bouzy, Marc M{\'e}tivier\\
Laboratoire d'Informatique de Paris Descartes  \\
Universit{\'e} Paris Descartes \\
45, rue des Saints P{\`e}res, 75006 Paris\\
\texttt{\small \{damien.pellier, bruno.bouzy, marc.metivier\}@parisdescartes.fr}
}

\date{June 2010}
\begin{document}
\maketitle

\begin{abstract}
In this paper, we introduce a new heuristic search algorithm based on mean values for real-time planning, called MHSP. It consists in associating the principles of UCT, a bandit-based algorithm which gave very good results in computer games, and especially in Computer Go, with heuristic search in order to obtain a real-time planner in the context of classical planning.  MHSP is evaluated on different planning problems and compared to existing algorithms performing on-line search and learning. Besides, our results highlight the capacity of MHSP to return plans in a real-time manner which tend to an optimal plan over the time which is faster and of better quality compared to existing algorithms in the literature.
\end{abstract}

\section{Introduction}

The starting point of this work is to apply UCT \cite{kocsis:06}, an efficient algorithm well-known in the machine learning and computer games communities, and originally designed for planning, on classical planning problems. UCT is designed for MDP, and based on bandit decision making \cite{auer:02}. In the background of the current paper that stresses time constraints, the interesting feature of UCT is its anytime property in a strong meaning. At any time, UCT is able to return the first action of a plan, a partial plan, a solution plan, or an optimal plan according to the given time. However, \cite{kocsis:06} did not give known successful applications in the classical planning domain yet \cite{ghallab:04}. Instead, UCT gave tremendous results in computer games, and specifically in Computer Go with the Go playing program Mogo \cite{gelly:06}. In Computer Go, UCT is efficient because the Go complexity is high, and because the Go games are played in real time, which fits the anytime property of UCT. Therefore, this paper focuses on how to give value to UCT-like algorithms in a sub-field of planning dealing with time constraints.

The state of the art of planning is huge \cite{roberts:09}, and we roughly divide it into two categories, off-line and on-line planning. In the context of off-line planning, the planners build solution plans, and then execute the actions of the plan. An important aspect is the existence of very good heuristic functions that drive the search efficiently toward the goal. The heuristic functions are built with the help of a planning graph \cite{blum:97}. The state-of-the-art planners use variants of depth first search, i.e., Enforced Hill Climbing \cite{hoffmann:01}, and may find a solution plan very quickly from the initial state to the goal state. However, although they find solution plans very quickly on sufficiently easy problems, these planners are not real-time planners, and they may fall if they have not enough time to find a solution plan.

Conversely, in the context of on-line planning, the planners make their decision in constant time, and then execute the corresponding action, or sequence of actions, in the world. The literature distinguishes two approaches, one based on MDP applied to non-deterministic problems, e.g., \cite{barto:95,hansen:01} and the other based on Real-Time Search (RTS), e.g., \cite{korf:90}. If the first approach was recently applied to planning \cite{fabiani:07}, the second approach RTS has been strongly linked, since the pioneering Korf's work on puzzles, to the development of video games in which the agents need good path-finding algorithms running in real time. This last approach was not broadened to the classical planning problems. Classically, there are several real-time searches, e.g., mini-min \cite{korf:88}, $\gamma$-Trap \cite{bulitko:04}, LRTS \cite{bulitko:06} or even A* \cite{hart:68}. These algorithms perform action selection by using heuristic search. Since the action selection time is limited, these algorithms explore a small part of the state space around the current state. The plans executed by the agent have no reason to be optimal. The RTS literature is concerned with convergence of these plans toward an optimal solution when the task is repeated iteratively. This opens the RTS literature mainly to learning. Then, considering the learning aspect as its main objective, and focused on convergence proofs, the literature reduced the action selection stage to a depth-one greedy search.

The aim of this paper is to focus on the action selection stage of RTS, to present MHSP, an heuristic search algorithm based on mean values, i.e., an adaptation of UCT to RTS, adapted in the context of classical planning. We show that MHSP performs better for action selection in the RTS background, with or without learning, than the existing real-time action selectors.

The outline of the paper is the following. Section \ref{sectionRealTimeSearch} describes the domain of Real-Time Search. Section \ref{sectionUCT} describes UCT principles. Section \ref{sectionMHSP} presents our new algorithm MHSP, an adaptation of UCT for RTS. Section \ref{sectionExperimentalResults} shows the experiments performed to prove the relevance of our approach. Section \ref{sectionConclusion} concludes and discusses future lines of research.

\section{Real-Time Search}
\label{sectionRealTimeSearch}

RTS is considered in the context of agent-centered search. While classical off-line search methods first find a solution plan, and then make the agent execute the plan, agent-centered search interleaves planning and execution. The agent repeatedly executes a task, called an episode or a trial. At each step, the agent is situated in a current state, and performs a search within the space around the current state in order to select his best action. The states encountered during the action selection stage are called the explored states in the following. The feature of RTS is that this search is performed in constant-time. When the action is selected, the agent executes the action, and reaches a new state. When the agent reaches the goal state, another trial or episode is launched, and so on. RTS can be considered with or without learning. Without learning, the efficiency of the agent is based on the ability of the search to select an action. With learning, the agent updates the heuristic value of the encountered states when some conditions happen, and the efficiency of the agent increases over repetitions. In the following, the states encountered by the agent are called the visited states.

The fundamental paper of RTS is Real-Time Heuristic Search \cite{korf:90}. Real-Time A* (RTA*) is an algorithm that computes a tree like A* does, but in constant time. When the time is elapsed, RTA* provides the first move of its current best plan, and executes it to reach a new node. From this new node, RTA* computes a tree again, executes the first move of the new current best plan, and so on until a goal node is reached. RTA* was designed in the spirit of two-player game programs that must play their moves in limited time. Like A*, RTA* uses an heuristic function. RTA* always finds a solution plan, even if not optimal. The learning version of RTA* is called LRTA*. When the heuristic value of a node is too low compared to the minimum value of its neighbors nodes, LRTA* updates the heuristic value of this node with the minimum value of its neighbors nodes. To this extent the heuristic function is modified, and learnt. When launched several times on the same problem, LRTA* is proved to converge to the optimal plan.

SLA* (Search and Learn A*) \cite{shue:93} is presented as an enhancement of LRTA*. SLA* includes a backtracking mechanism when an update happens. As LRTA* does, SLA* updates the heuristic value of a node. Additionally, when an update occurs, SLA* iteratively updates the neighboring nodes of the updated node as well. Actually, in one trial, SLA* learns the heuristic function, and finds the optimal solution. However, since the first trial can be very long, SLA* cannot be used in practice.

In \cite{furcy:00}, Furcy presents FALCONS, a learning algorithm that converges quickly under some assumptions. Its main feature is to compute two heuristic functions, one for each way from start to goal, and from goal to start.

In \cite{shimbo:03a}, Shimbo shows how weighted A* and upper bound search are worth considering in real-time search. In weighted A*, the heuristic function has a weight $1+\epsilon$. The greater $\epsilon$ the most important the heuristic function. The risk is that the heuristic function becomes non admissible. Nevertheless, even with non admissible functions, the heuristic search finds good plans, although not optimal. Upper bound search limits the search for path with length inferior to the upper bound. Weighted A* and upper bound search are sub-optimal.

$\gamma$-Trap \cite{bulitko:04} includes some lookahead to smooth the bad effects of the heuristic function. When compared to LRTA* and FALCONS, because it selects a sequence of actions instead the first action, $\gamma$-Trap yields improvements of 5 to 30 folds in convergence speed. $\gamma$ is an optimality weight associated to the cost function $G$ (with $0<\gamma<=1$). $\gamma$ has the same purpose as $\epsilon$ in weighted A*: balancing the weights of $G$ and $H$.

LRTS \cite{bulitko:06} is a unifying framework for learning in real-time search. It includes LRTA*, SLA* and $\gamma$-Trap. LRTS features are: learning in real-time (inherited from LRTA*), lookahead (inherited from $\gamma$-Trap, and the Korf's work), backtracking (inherited from SLA*), and weighted search (inherited from $\gamma$-Trap and weighted A*).

Finally, Bulitko \cite{bulitko:08} describes dynamic control in real-time heuristic search.

To sum up, the limitation of RTS is to focus on the learning part and less on the search for action selection. \cite{korf:90} and following works prove the convergence of their learning real-time algorithms given that the action selection is a depth-one greedy search. However, the Korf's work shows the importance of an efficient search for action selection in other papers. The exception to this limitation is $\gamma$-Trap and LRTS in which lookahead is used for action selection. The $\gamma$-Trap lookahead is a kind of breadth first search. Consequently, it is of interest to see if an adaptation of UCT can be used efficiently for action selection in RTS. Then, this adaptation will be used with or without learning.

\section{UCT} \label{sectionUCT}

UCT worked well in Go playing programs, and it was used under many versions leading to the Monte-Carlo Tree Search (MCTS) framework \cite{chaslot:08}. While time remains, a MCTS algorithm iteratively grows up a tree in the computer memory by following the steps below:
\begin{enumerate}
\item Starting from the root, browse the tree until reaching a leaf by using (\ref{eq:UCBselect}).
\item Expand the leaf with its child nodes.
\item Choose one child node.
\item Perform a random simulation starting from this child node until the end of the game, and get the return, i.e., the game's outcome.
\item Update the mean value of the browsed nodes with this return.
\end{enumerate}

With infinite time, the root value converges to the mini-max value of the game tree \cite{kocsis:06}. The UCB selection rule (\ref{eq:UCBselect}) answers the requirement of being optimistic when a decision must be made facing uncertainty \cite{auer:02}.
\begin{equation}
N_{select} = \text{arg}\max_{n\in N}\{m + C\sqrt{\frac{\log{p}}{s}} \}
\label{eq:UCBselect}
\end{equation}
$N_{select}$ is the selected node, $N$ is the set of children, $m$ is the mean value of node $n$, $s$ is the number of iterations going through $n$, $p$ is the number of iterations going through the parent of $n$, and $C$ is a constant value setup experimentally. Equation (\ref{eq:UCBselect}) uses the sum of two terms: the mean value $m$, and the UCB bias value which guarantees exploration. The respect of the UCB selection rule  guarantees the completeness and the correctness of the algorithm.

\section{MHSP} \label{sectionMHSP}

This section defines our algorithm MHSP (Mean-based Heuristic Search for real-time Planning). We made two important choices in designing MHSP after which we give the pseudo-code of MHSP.

\subsection{Heuristic values replace simulation}

On planning problems, random simulations are not appropriate. Browsing randomly the state space does not enable the algorithm to reach goal states sufficiently often. Many runs complete without reaching goal states. Therefore, replacing the simulations by a call to the heuristic becomes mandatory.

In Computer Go, the random simulations were adequate mainly because they always completed after a limited number of moves, and the return values (won or lost) were roughly equally distributed on most positions of a game. Furthermore, the two return values correspond to actual values of a completed game. In planning, one return means that a solution has been found (episode completed), and the other not. This simulation difference is fundamental between the planning problem, and the two-player game playing problem.

In planning, the heuristic values bring appropriate knowledge into the returns. Consequently, using heuristic values in MHSP should be positive bound to the condition that the heuristic value generator is good, which is the case in planning \cite{blum:97}. In Computer Go, replacing the simulations by evaluation function calls is forbidden by fifty years of computer Go history which experienced the converse path: the complex evaluation functions have been replaced by pseudo-random simulations.

To sum up, in MHSP, we replace stage (4) of MCTS above by a call to an heuristic function.

\subsection{Optimistic initial mean values}

Computer games practice shows that the UCB bias of (\ref{eq:UCBselect}) can merely be removed, provided the mean values of nodes are initialized with sufficiently optimistic values. This simplification removes the problem of tuning $C$. Generally, to estimate a given node, the planning heuristics give a path length estimation. Convergence to the best plan is provided by admissible heuristics, i.e., optimistic heuristics. Consequently, the value returned by planning heuristics on a node can be used to initialize the mean value of this node.

In MHSP, the returns are negative or zero, and they must be in the opposite of the distance from $s$ to $g$. Thus, we initialize the mean value of a node with $\Delta(s,g)$ which is minus the distance estimation to reach $g$ from $s$. With this initialization policy, the best node according to the heuristic value will be explored first. Its value will be lowered after some iterations whatever its goodness, and then the other nodes will be explored in the order given by the heuristic.

\subsection{The algorithm}

\LinesNumbered
\DontPrintSemicolon
\begin{algorithm}[!t]
\SetKwData{Resolvers}{resolvers}
\SetKwFunction{Select}{Select}
\SetKwFunction{heuristicExpand}{Expand}
\caption{MHSP($O,s_0,g$)}
\BlankLine
$C[s_0] \leftarrow \emptyset$ ; $R[s_0] \leftarrow \Delta(s_0,g)$ ; $V[s_0] \leftarrow 1$; $\pi \leftarrow nil$ \;
\While{$has\_time$}{
  $s \leftarrow s_0$ \;
  \While{$g \not\subseteq s$ and $V[s] \not=$ 1} {
    $s \leftarrow argmax_{s' \in C[s]}(R[s']/V[s'])$ \;
  }
  $reward \leftarrow (R[s_0]/V[s_0]) + 1$\;
  \If{$g \subseteq s$}{
    $reward \leftarrow 0$\;
  }\uElseIf{$V[s] = 1$}{
    $A \leftarrow \{ a \ | \ a$ ground instance of an operator in $O$ and $\text{\it precond}(a) \subseteq s\}$\;
    \ForEach{$a \in A$}{
      $s' \leftarrow (s \cup \text{\it effects}^{+}(a)) - \text{\it effects}^{-}(a)$ \;
      $C[s'] \leftarrow C[s] \cup \{s'\}$ \;
      $R[s'] \leftarrow \Delta(s',g)$ \;
      $P[s'] \leftarrow s$ \;
      $V[s'] \leftarrow 1$ \;
    }
    \If{$C[s] \not= \emptyset$}{
      $s \leftarrow argmax_{s' \in C[s]}(R[s'])$\;
      $reward \leftarrow R[s]$\;
    }
  }
  $i \leftarrow 0$ \;
  \While{$s \not= s_0$}{
    $s \leftarrow P[s]$ \;
    $R[s] \leftarrow R[s] + (reward - i)$ \;
    $V[s] \leftarrow V[s] + 1$ \;
    $i \leftarrow i + 1$ \;
  }
  \If{$g \subseteq s$}{
    $\pi' \leftarrow \text{\it reconstruct\_solution\_plan}()$ \;
    \lIf{\textit{length}($\pi$) $>$ \textit{length}($\pi'$)}{
      $\pi \leftarrow \pi'$\;
    }
  }
}
\lIf{$\pi = nil$ }{
  \Return $\text{\it reconstruct\_best\_plan}()$ \;
}\lElse{
  \Return $\pi$ \;
}

\label{Algo:UCTP}
\end{algorithm}

MHSP algorithm is shown in Algo.~1 : $O$ is the set of operators, $s_0$ the initial state, $g$ the goal, $C[s]$ the set of children of state $s$, $R[s]$ the cumulative return of state $s$, $V[s]$ the number of visits of state $s$, and $P[s]$ the parent of $s$.

The outer $while$ (line 2) ensures the real-time property. The first inner $while$ (line 4) corresponds to stage (1) in MCTS. The default reward is pessimistic: $(R[s_0]/V[s_0])+1$ is the current pessimism threshold. The first two $if$ test whether the inner $while$ has ended up with a goal achieved (line 7) or with a leaf (line 9). If the goal is not reached, the leaf is expanded, stage (2) in MCTS. The second $if$ corresponds to stage (3). Stage (4) is performed by writing $\Delta(s',g)$ into the return. The second inner $while$ (line 22) corresponds to stage (5).

Function $\text{\it reconstruct\_solution\_plan}()$ browses the tree by selecting the child node with the best mean, which produces the solution plan. Function $\text{\it reconstruct\_best\_plan}()$ browses the tree by selecting the child node with the best number of visits. The best plan reconstruction happens when the time is over before a solution plan has been found. In this case, it is important to reconstruct a robust plan, may be not the best one in terms of mean value. With the child with the best mean, a plan with newly created nodes could be selected, and the plan would not be robust. Conversely selecting the child with the best number of visits ensures that the plan has been tried many times, and should be robust to this extent.

\section{Experiments}
\label{sectionExperimentalResults}

The aim of the experiments described in this section is to show that MHSP is better than existing correct and complete algorithms at action selection in the background of RTS, used with or without learning, for different decision times, on a set of planning problems.

\subsection{Planners and domains}
\label{subSectionPlannersAndDomains}

We compare MHSP with two algorithms: A* and Breadth-First Search (BFS). We chose A* because it is the reference algorithm in planning (LRTA*). It is a best-first algorithm that aims at minimizing the classical heuristic function $f$ of A*. A* expands nodes in the open list with $f$ values decreasing with the running time, the $f$ value reaching zero with sufficient time. To this extent, A* can be stopped at anytime. When A* is stopped, the path from the last expanded node to the root node gives the ``best'' action selected by A*. Beside, we chose BFS. BFS is a simple generalization of current action selectors, such as LRTS or $\gamma$-Trap, used in RTS for path-finding to other planning domains. We did not choose Depth-First algorithms, such as Mini-min, since they hardly fit the real-time constraint. Finally, our three algorithms are MHSP, A*, and BFS.

As mentioned in introduction, RTS algorithms are mainly applied to path-finding for video games. In order to extend the existing test domain, we selected other domains and problems from International Planning Competition\footnote{For a description and formalization of these benchmark domains and problems, see the official page of IPC.}, which illustrates the effectiveness of our techniques implemented in MHSP. The domains are ferry, gripper, and satellite. For each domain, we selected 20 problems ranked by complexity in terms of fact number and operator number. In the rest of the paper, we just show results from this set of problems in order to illustrate the power of MHSP.

\subsection{Settings} \label{subSectionSettings}

We designed four tests to underline the effectiveness of MHSP. Test 1 is global, and does not especially focus on action selection: it gives the average length of solution plans found by the three algorithms for different decision times, and representative problems. It intends to show that MHSP is globally better than A* and BFS in terms of solution plan length. Test 1 does not includes learning.

Test 2 re-perform test 1 with learning by using the update rule (\ref{eq:UpdateH}) on nodes $s$ visited during the episodes. Rule (\ref{eq:UpdateH}) is not applied on nodes explored during action selection time. Test 2 intends to show the consequences of the three action selectors on the convergence of solution plans toward optimal plans when the episode number increases.

\begin{equation}
H(s) = max \{ H(s), min_{s' \in C[s]} \{ 1 + H(s') \} \}
\label{eq:UpdateH}
\end{equation}

Test 3 is the most important test to underline the ability of the algorithms to performs effective action selection, or partial plan selection in real time. This test makes the decision time vary, and assesses the quality of partial plans obtained. The quality of partial plans is estimated with two distances: the distance to the goal (or goal distance) and the distance to the optimum.

The distance to the goal of a partial plan is the length of the optimal plan linking the end state of this partial plan to the goal state. When the distance to the goal diminishes, the partial plan has been built in the appropriate direction. When the distance to the goal is zero, the partial plan is a solution plan.

The distance to the optimum of a partial plan is the length of the partial plan, plus the distance to the goal of the partial plan, minus the length of the optimal plan. When the distance to the optimum of a partial plan is zero, the partial plan is the beginning of an optimal plan. The distance to the optimum of a solution plan is the difference between its length and the optimal length. The distance to the optimum of the void plan is zero. The distance to the goal and the distance to the optimal plan is zero. Conversely, when the distance to the goal and the distance to the optimum of a partial plan are zero, the partial plan is an optimal plan. For each problem, the results are shown with figures giving the distance to the goal and the distance to the optimum of the partial plan in the running time. These distances are computed by calling an optimal planner.


Finally, all the tests were conducted on an Intel Core 2 Quad 6600 (2.4Ghz) with 4 Gbytes of RAM. The implementation of MHSP used for experiments is written in Java based on the PDDL4J library\footnote{\url{http://sourceforge.net/projects/pdd4j/}}.

\subsection{Test 1}
\label{subSectionTest1}

\rowcolors{3}{white}{lightgray}
\begin{table*}[t]
\begin{center}
\begin{tabular}{llccccccc}
\hline
\textbf{problem} & \textbf{algo.} & \textbf{decision} & \textbf{avr.}  & \textbf{avr.} & \textbf{opt.} & \textbf{max}  & \textbf{min}  & \textbf{failure} \\
 & & & \textbf{time}  & \textbf{length} & \textbf{length} & \textbf{length}  & \textbf{length}  & \textbf{\%} \\
\hline
\hline
ferry-05     & A*   & 40   & 1.09    & 26.26  & 18 & 277 & 19 & 0   \\
ferry-05     & BFS  & 40   & 8.91    & 27.76  & 18 & 567 & 18 & 16  \\
ferry-05     & MHSP & 40   & 0.59    & 18.02  & 18 & 19  & 18 & 0   \\
ferry-10     & A*   & 200  & 97.9    & 184.94 & 35 & 807 & 42 & 32  \\
ferry-10     & BFS  & 200  & 8.91    & 27.76  & 35 & 109 & 35 & 0   \\
ferry-10     & MHSP & 200  & 0.59    & 18.02  & 35 & 36  & 35 & 0   \\
ferry-15     & A*   & 2000 & 157.85  & 31.95  & 51 & 88  & 58 & 55  \\
ferry-15     & BFS  & 2000 & 103.75  & 51.45  & 51 & 52  & 51 & 0   \\
ferry-15     & MHSP & 2000 & 86.45   & 52.45  & 51 & 53  & 51 & 0   \\
ferry-20     & A*   & 4000 & --      & --     & 73 & --  & -- & 100 \\
ferry-20     & BFS  & 4000 & --      & --     & 73 & --  & -- & 100 \\
ferry-20     & MHSP & 4000 & 260.49  & 74.87  & 73 & 78  & 73 & 0   \\
gripper-05   & A*   & 50   & 0.56    & 15.04  & 15 & 17  & 15 & 0   \\
gripper-05   & BFS  & 50   & 0.71    & 15     & 15 & 15  & 15 & 0   \\
gripper-05   & MHSP & 50   & 0.49    & 15     & 15 & 15  & 15 & 0   \\
gripper-10   & A*   & 165  & 92.28   & 140.04 & 29 & 651 & 33 & 36  \\
gripper-10   & BFS  & 165  & 7.86    & 37     & 29 & 37  & 37 & 0   \\
gripper-10   & MHSP & 165  & 5.08    & 29     & 29 & 29  & 29 & 0   \\
gripper-15   & A*   & 450  & 160.1   & 47.54  & 45 & 229 & 77 & 70  \\
gripper-15   & BFS  & 450  & 31.42   & 46.52  & 45 & 47  & 45 & 0   \\
gripper-15   & MHSP & 450  & 38.53   & 54.88  & 45 & 55  & 53 & 0   \\
gripper-20   & A*   & 1100 & --      & --     & 59 & --  & -- & 100 \\
gripper-20   & BFS  & 1100 & 102.76  & 61     & 59 & 61  & 61 & 0   \\
gripper-20   & MHSP & 1100 & 134.86  & 73.92  & 59 & 75  & 73 & 0   \\
satellite-05 & A*   & 300  & 3.49    & 15.08  & 15 & 18  & 15 & 0   \\
satellite-05 & BFS  & 300  & 20.28   & 63.72  & 15 & 522 & 19 & 0   \\
satellite-05 & MHSP & 300  & 3.01    & 15     & 15 & 15  & 15 & 0   \\
satellite-10 & A*   & 2000 & --      & --     & 29 & --  & -- & 100 \\
satellite-10 & BFS  & 2000 & --      & --     & 29 & --  & -- & 100 \\
satellite-10 & MHSP & 2000 & 67.912  & 31.0   & 29 & 31  & 31 & 0   \\
\hline
\end{tabular}
\end{center}
\caption{Results of the test 1 on ferry, gripper and satellite domains without learning}
\label{Tab:Test1}
\end{table*}

Table \ref{Tab:Test1} takes the following inputs: the domain (ferry, gripper or satellite), the problem number, the algorithm used for action selection, the decision time. The outputs are: the optimal plan length, the average time spent for one episode, the average solution plan length, the maximal solution plan length, and the minimal plan length found by the algorithm, and the percentage of failures.

The optimal plan length, computed off-line, is used as a reference. On ferry-05 with a decision time of 40ms, MHSP executes solution plans that are almost optimal (18.02 against 18) while A* and BFS are far from optimal (16.26 and 27.76). The maximal plan length is very high for A* and BFS (277 and 567) and almost optimal for MHSP (19). The minimal plan length is the optimal one for BFS and MHSP. In order to limit the time of the experiments, there is a maximal episode number (50). Consequently, an algorithm that does not reach the goal during an episode failed. Here, BFS has 16\% of failure rate on the 50 episodes. Furthermore, MHSP is the fastest algorithm. In the beginning of an episode, all the algorithms use the total time to decide. However, when the episode reaches its end, the action selection is easier than before, and some algorithms do not use all the available time, and are faster than other. MHSP is clearly the fastest because it finds the goal more easily than A* or BFS when the goal is not far.

On ferry-10 with a decision time of 200ms, MHSP executes solution plans that are almost optimal (35.66 against 35) while A* and BFS are again far from optimal (181.94 and 41.34). The maximal plan length is very high for A* and BFS (807 and 109) and almost optimal for MHSP (36). The minimal plan length is the optimal one for BFS and MHSP. Here, BFS has 32\% of failure rate.

On gripper-05 with a decision time of 50ms, MHSP, A* and BFS are almost optimal. On gripper-10 with a decision time of 165ms, MHSP executes optimal plans (29) while A* and BFS are again far from optimal (140.04 and 37). Here, A* has 36\% of failure rate. 

Finally, on satellite-05 with a decision time of 300ms, MHSP, A* and BFS are almost optimal with a slight preference for MHSP. Now on satellite-10 with a decision time of 2000ms, BFS and A* do not find any solution unlike MHSP. The main reason for this result is the branching factor of the problems which is greater than the other studied problems (e.g., 66 for satellite-10, 13 for gripper-20 and 16 for ferry-20). Thus, this high branching factor strongly penalizes the exhaustive search strategy of A* and BFS.

To sum up the first test, MHSP finds plans shorter than A* or BFS, and MHSP is faster than A* and BFS.

\subsection{Test 2} \label{subSectionTest2}

\rowcolors{3}{white}{lightgray}
\begin{table*}[!t]
\begin{center}
\begin{tabular}{llcccccccc}
\hline
\textbf{problem} & \textbf{algo.} & \textbf{decision} & \textbf{avr.}  & \textbf{avr.} & \textbf{opt.} & \textbf{max}  & \textbf{min}  & \textbf{failure} \\
\textbf{} & \textbf{} & \textbf{} & \textbf{time}  & \textbf{length} & \textbf{length} & \textbf{length}  & \textbf{length}  & \textbf{\%} \\
\hline
\hline
ferry-05     & A*   & 40   & 0.68    & 19.22  & 18 & 31  & 18  & 0   \\
ferry-05     & BFS  & 40   & 12.03   & 133.18 & 18 & 532 & 18  & 14  \\
ferry-05     & MHSP & 40   & 0.48    & 18.02  & 18 & 19  & 18  & 0   \\
ferry-10     & A*   & 200  & 9.36    & 48.12  & 35 & 67  & 37  & 0   \\
ferry-10     & BFS  & 200  & 17.22   & 78.5   & 35 & 892 & 35  & 0   \\
ferry-10     & MHSP & 200  & 6.24    & 35     & 35 & 35  & 35  & 0   \\
ferry-15     & A*   & 2000 & 112.03  & 62.36  & 51 & 81  & 57  & 55  \\
ferry-15     & BFS  & 2000 & 104.31  & 51.75  & 51 & 53  & 51  & 0   \\
ferry-15     & MHSP & 2000 & 87.39   & 52.8   & 51 & 53  & 51  & 0   \\
ferry-20     & A*   & 4000 & 247.32  & 121.1  & 73 & 146 & 107 & 0   \\
ferry-20     & BFS  & 4000 & 284.44  & 73.8   & 73 & 75  & 75  & 35  \\
ferry-20     & MHSP & 4000 & 255.31  & 73.6   & 73 & 73  & 73  & 15  \\
gripper-05   & A*   & 50   & 0.45    & 15     & 15 & 15  & 15  & 0   \\
gripper-05   & BFS  & 50   & 35.72   & 34.06  & 15 & 615 & 15  & 68  \\
gripper-05   & MHSP & 50   & 0.48    & 15     & 15 & 15  & 15  & 0   \\
gripper-10   & A*   & 165  & 76.48   & 35.04  & 29 & 43  & 29  & 0   \\
gripper-10   & BFS  & 165  & 9.96    & 32.22  & 29 & 37  & 29  & 0   \\
gripper-10   & MHSP & 165  & 4.8     & 29     & 29 & 29  & 29  & 0   \\
gripper-15   & A*   & 450  & 56.32   & 54.32  & 45 & 57  & 49  & 14  \\
gripper-15   & BFS  & 450  & 36.92   & 54.8   & 45 & 318 & 45  & 4   \\
gripper-15   & MHSP & 450  & 36.52   & 46.76  & 45 & 55  & 45  & 0   \\
gripper-20   & A*   & 1100 & --      & --     & 59 & --  & --  & 100 \\
gripper-20   & BFS  & 1100 & 132.44  & 74.12  & 59 & 75  & 73  & 2   \\
gripper-20   & MHSP & 1100 & 99.47   & 59.06  & 59 & 63  & 73  & 0   \\
satellite-05 & A*   & 300  & 3.54    & 15.62  & 15 & 18  & 15  & 0   \\
satellite-05 & BFS  & 300  & 6.1     & 18.98  & 15 & 41  & 19  & 0   \\
satellite-05 & MHSP & 300  & 3.17    & 15     & 15 & 15  & 15  & 0   \\
satellite-10 & A*   & 2000 & --      & --     & 29 & --  & --  & 100 \\
satellite-10 & BFS  & 2000 & --      & --     & 29 & --  & --  & 100 \\
satellite-10 & MHSP & 2000 & 65.612  & 31.2   & 29  & 32 & 30  & 0   \\
\hline
\end{tabular}
\end{center}
\caption{Results of the test 1 on ferry, gripper and satellite domains with learning}
\label{Tab:test2}
\end{table*}

Table~\ref{Tab:test2} shows the three algorithms performances when learning is applied. Compared to results of table~\ref{Tab:Test1}, we can observe that learning most often improves the quality of best plans. Indeed, except for BFS in gripper-20, the minimal plan length is always smaller with learning than without, if it was not already optimal in test 1.

Moreover learning enabled BFS and A* to find solution plans in ferry-15. None of them reached an optimal plan, but BFS' best plan is two actions far from it. It also enabled MHSP to become optimal in satellite-05.

\begin{figure}[!t]
\begin{center}
 \subfigure[ferry-10]{
   \includegraphics[scale=0.67]{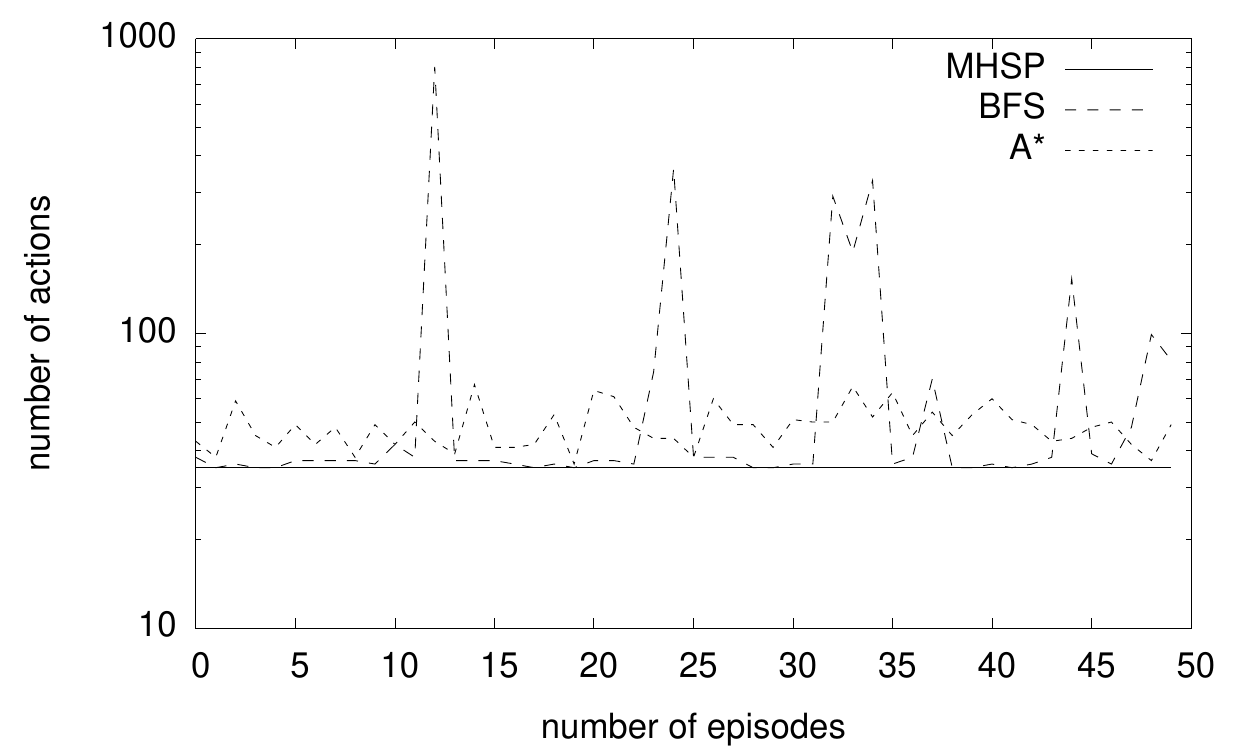}
   \label{Fig:FER-PB10-UCT-BFS-ASTAR-WL}
 }
 \subfigure[gripper-10]{
   \includegraphics[scale=0.67]{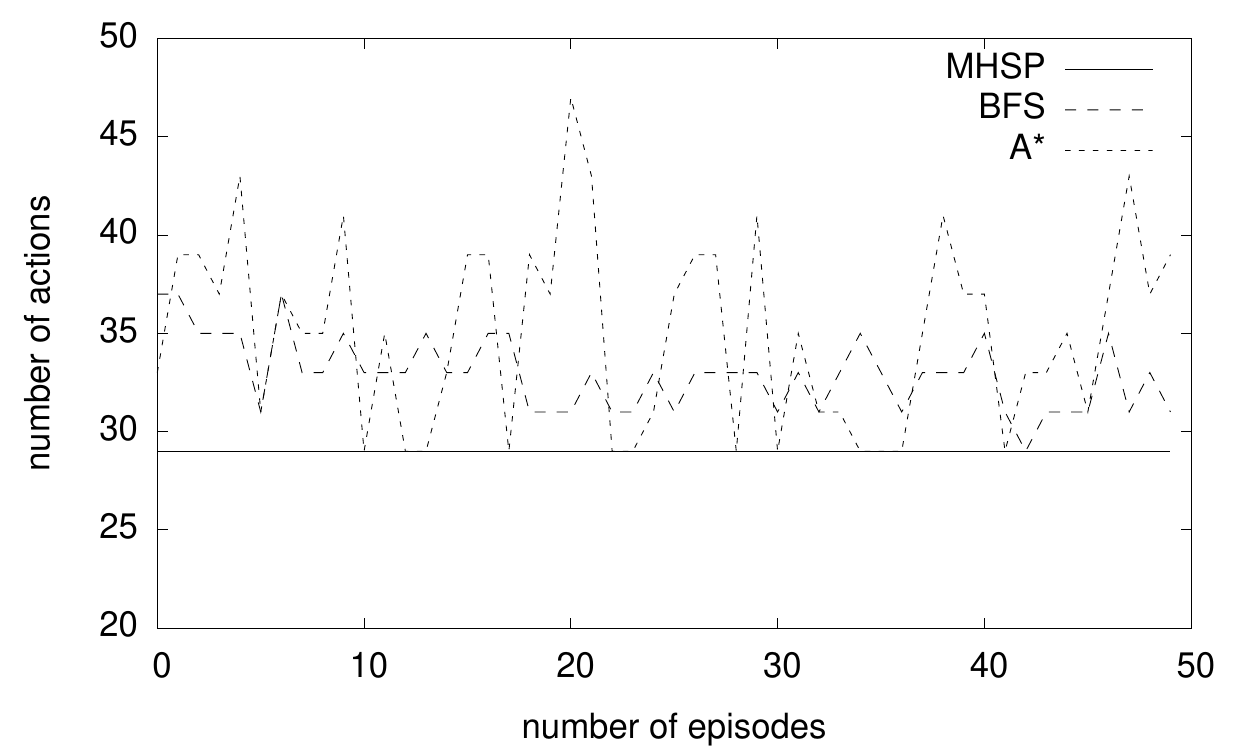}
   \label{Fig:GRI-PB10-UCT-BFS-ASTAR-WL}
 }
 \subfigure[satellite-05]{
   \includegraphics[scale=0.67]{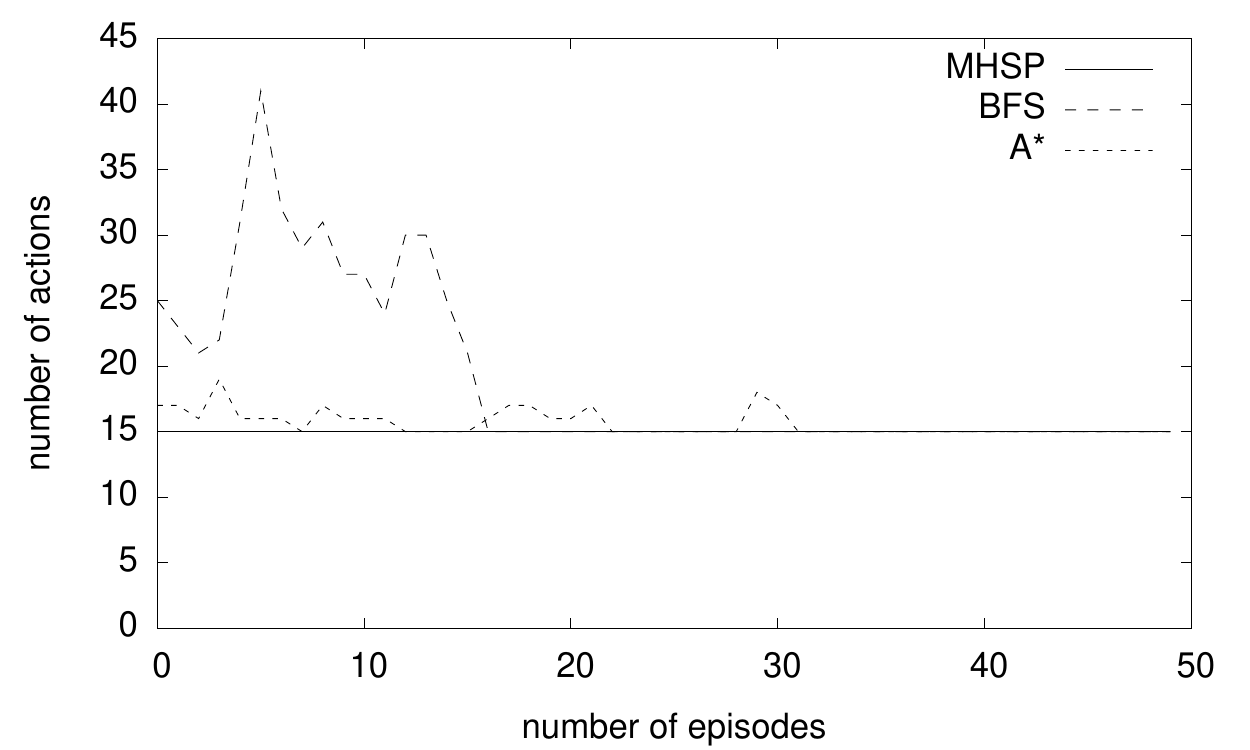}
   \label{Fig:SAT-PB05-UCT-BFS-ASTAR-WL}
 }
\end{center}
\caption{Test 2 \--- Convergence of the real-time planning algorithm with learning}
\label{Fig:Test2}
\end{figure}


In order to illustrate algorithms' behaviors over time, figure~\ref{Fig:Test2} shows the evolution of plan length according to episodes, in three problems : ferry-10, gripper-10 and Satellites-5. As we can see, on these problems, MHSP reaches an optimal plan very quickly and plan length is almost constant and stable. Conversely, A* and BFS are not optimal in ferry-10 and gripper-10 and are very unstable. The BFS peaks correspond to the maximal plan length allowed.

However, on these figures, learning is not observed by a clear decreasing plan length as expected. There are two reasons. First, the update rule is applied in the visited nodes, and not in the explored nodes. Therefore the action selection strategy does not really impact on the plan length when the episode number increases. Second, the algorithms use heuristic values computed off-line by planning graph techniques, that are almost optimal on problems with a sufficiently low complexity. Consequently, the update rule is not effective very often.


\subsection{Test 3}
\label{subSectionTest3}

\begin{figure}
\begin{center}
 \subfigure[gripper-05 \--- overview]{
   \includegraphics[scale=0.5]{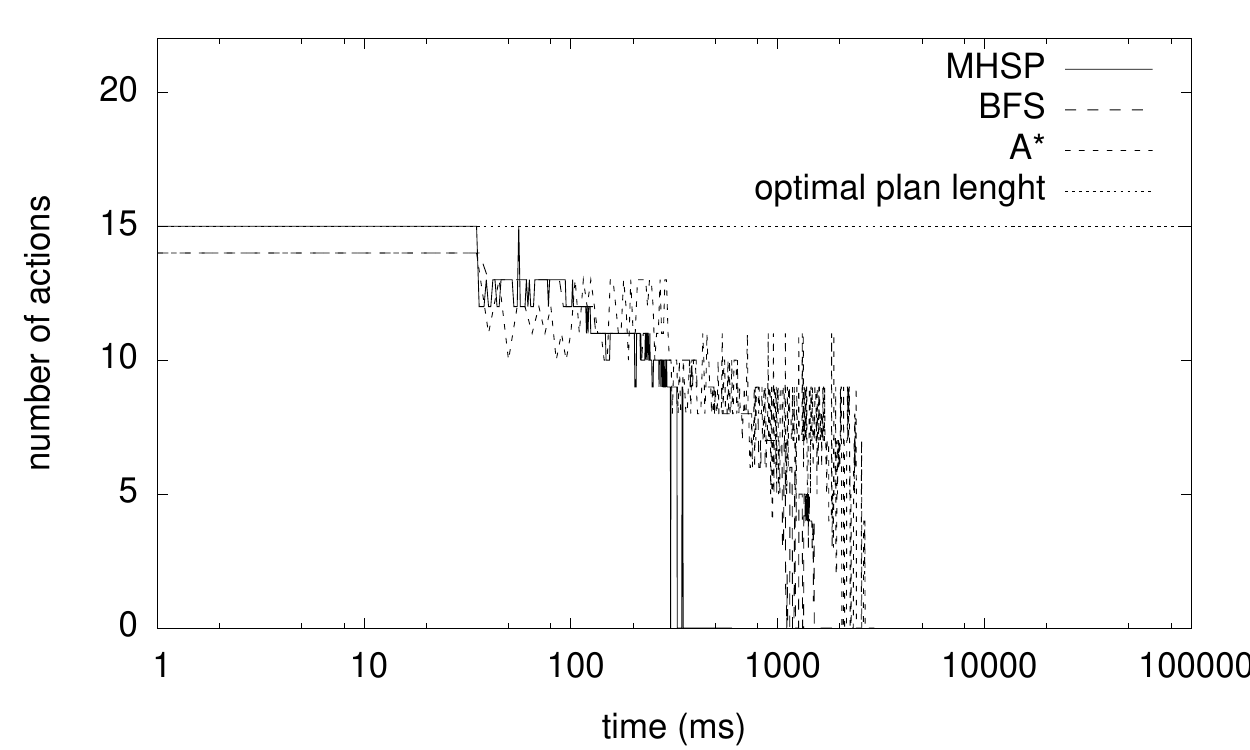}
   \label{Fig:GRI-PB05-UCT-BFS-ASTAR}
 }
 \subfigure[gripper-05 \--- A*]{
   \includegraphics[scale=0.5]{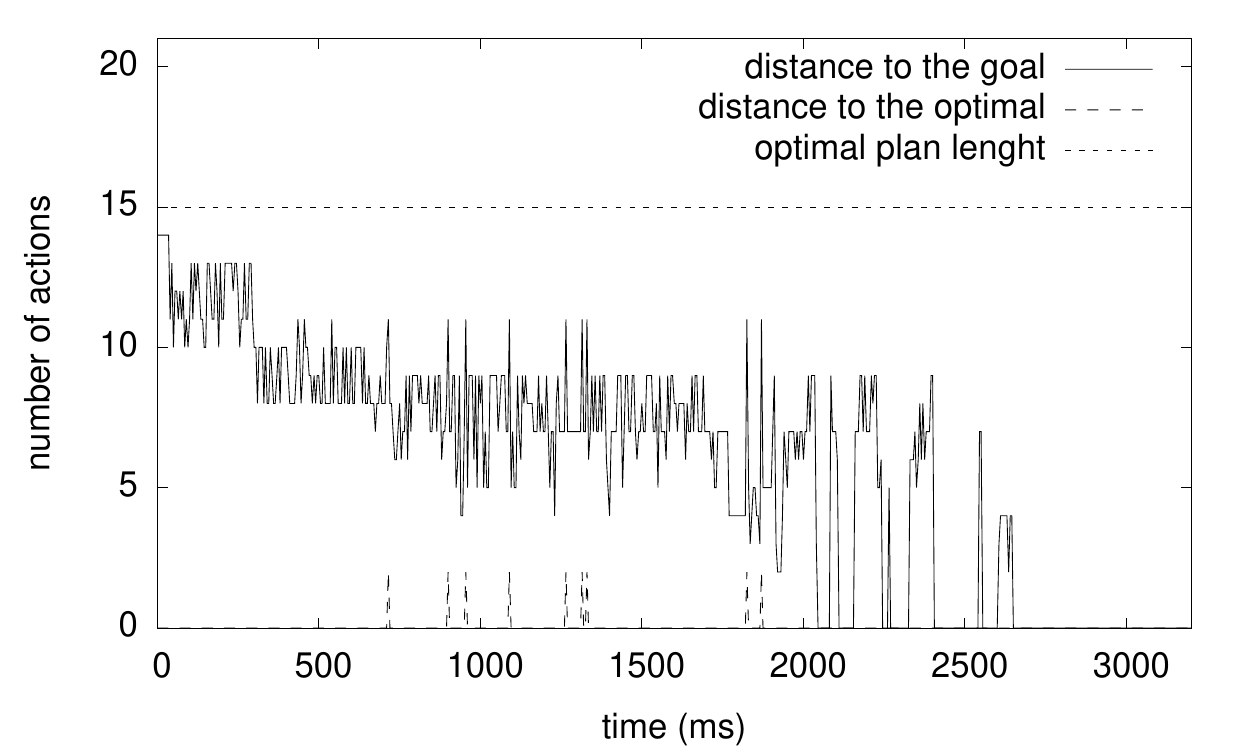}
   \label{Fig:GRI-ASTAR-PB05-S0-E3200-P5}
 }
 \subfigure[gripper-05 \--- BFS]{
   \includegraphics[scale=0.5]{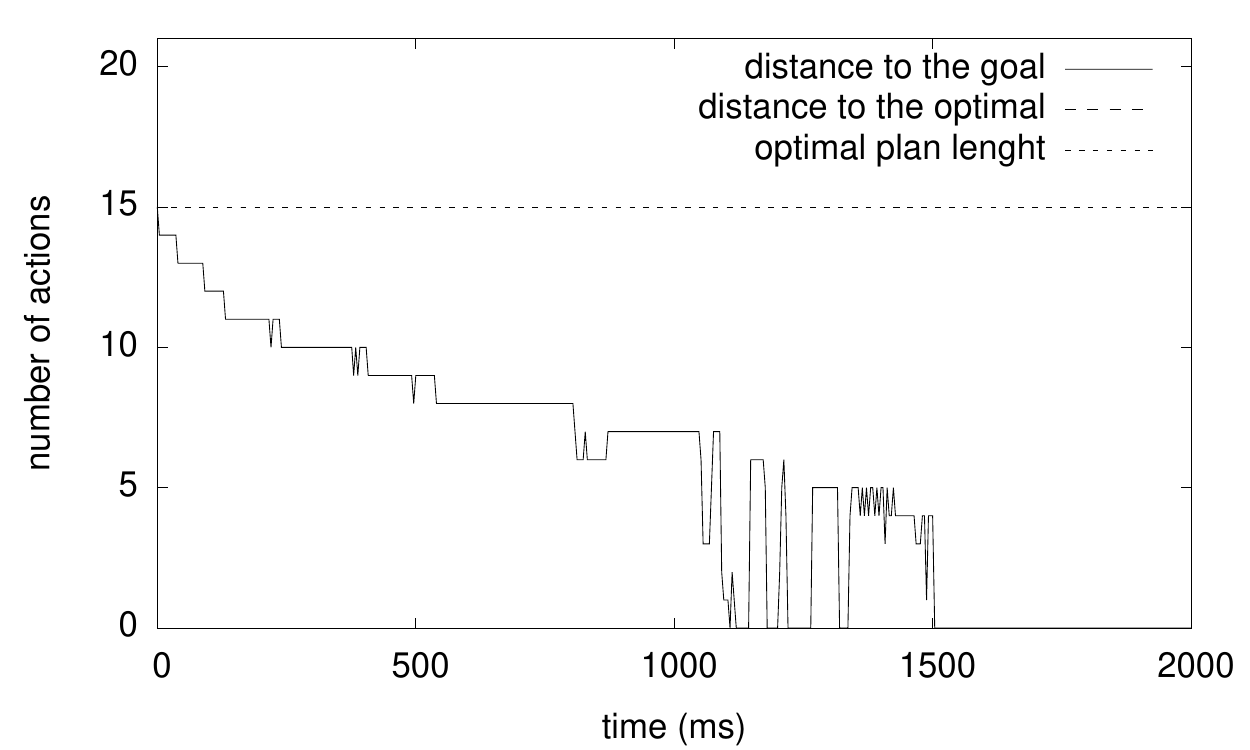}
   \label{Fig:GRI-BFS-PB05-S0-E2000-P4}
 }
 \subfigure[gripper-05 \--- MHSP]{
   \includegraphics[scale=0.5]{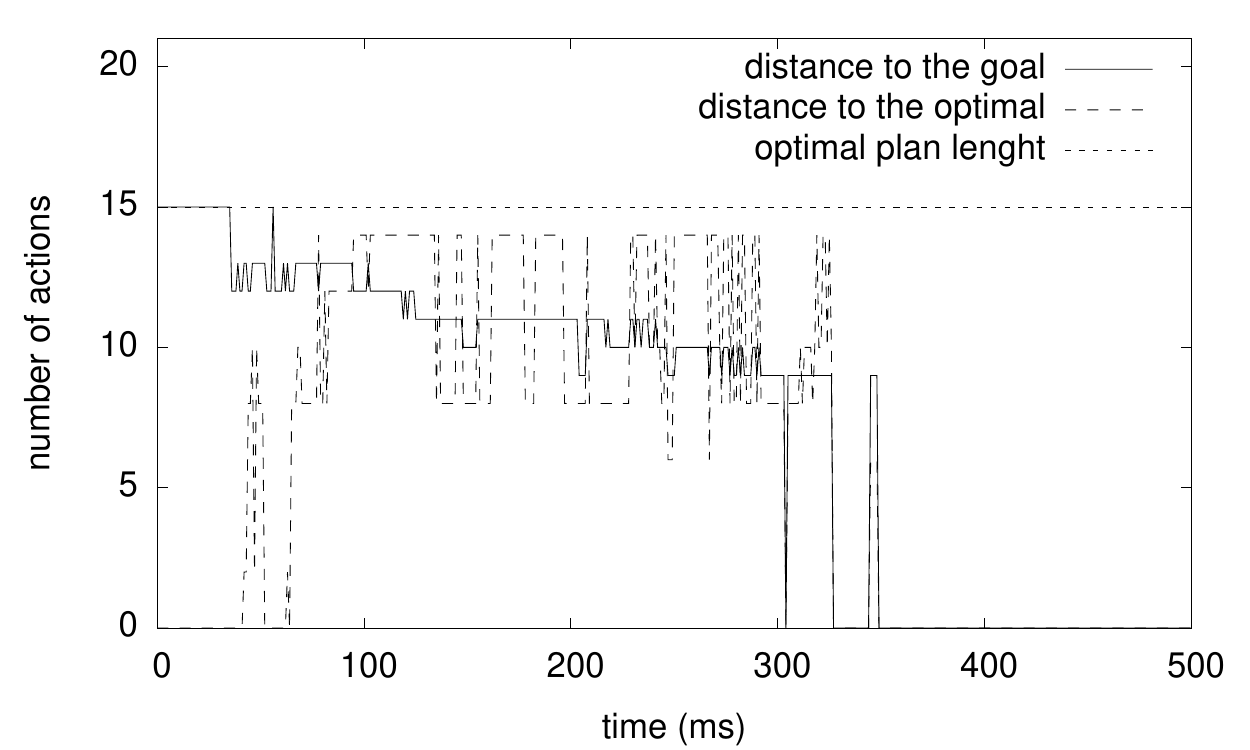}
   \label{Fig:GRI-UCT-PB05-S0-E600-P1}
 }
\end{center}
\caption{Test 3 \--- Quality of the action selection}
\label{Fig:Test3}
\end{figure}

Test 3 assesses the quality of partial plans available at the end of the action selection stage according to the time given to the decision. Figure~\ref{Fig:Test3}b, c and d show the distance to the goal, and the optimal plan distance, for each algorithm according to decision time on problem gripper-05. Figure~\ref{Fig:Test3}a gives an overview of the distance to the goal of the three algorithms in log-scale decision time.

In these results, we observe that A* needs a decision time of at least 2650ms to always find the optimal plan, while BFS needs at least 1504ms and MHSP only 349ms. Whatever the given decision time, A* provides partial plan near to the optimal. BFS provides only partial plans of optimal plans. Finally, during its search, MHSP provides partial plans having high optimal plan distance. This can be explained by the end of the partial plans that are unstable. Selecting shorter partial plans would result in a better optimal plan distance.

Table~\ref{Tab:Test3} sums up these results and adds results from ferry-05 and satellite-05. As in gripper, we can see that MHSP is the fastest algorithm to finds optimal plans.

\rowcolors{2}{white}{lightgray}
\begin{table}[h!]
\begin{center}
\begin{tabular}{llr}
\hline
\textbf{problem} & \textbf{algorithm} & \textbf{time (ms)}\\
\hline
\hline
ferry-05     & A*   & 808   \\
    & BFS  & -   \\
   & MHSP & 288   \\
\hline
gripper-05   & A*   & 1650   \\
  & BFS  & 1504   \\
  & MHSP & 349   \\
\hline
satellite-05 & A*   & 8180  \\
& BFS  & - \\
& MHSP & 1710  \\
\hline
\end{tabular}
\end{center}
\caption{Time to find a solution plan on ferry, gripper and satellite domain}
\label{Tab:Test3}
\end{table}


\section{Discussion and future works}
\label{sectionConclusion}

In this paper, we presented and study a new heuristic search algorithm based on mean values for real-time search (RTS), called MHSP, adapted in the context of classical planning. This algorithm combines an heuristic search and the learning principles of UCT algorithm, i.e., states' values based on mean returns, and optimism in front of uncertainty.

MHSP computes mean values for decision and not direct values. It means that the value of an action depends on every nodes explored beneath that action, and not only on the best node found. This fact may have a strong impact on the way the system explores nodes because the system may focus on action permitting to reach globally good node, and not on the action enabling to reach the node with the best heuristic value. In a time constrained context, focusing on action which leads to globally good nodes instead of just one node may limit the effect of strongly erroneous heuristic values. It enables to subtly explore the tree. The more complex the problem is, the more visible should be this effect.

Three tests were designed in order to compare MHSP, A* and BFS in RTS. The first one gave an overview of the global effectiveness of each algorithm to find good plans in different problems from ferry, gripper and satellite domains. It showed that MHSP is globally better than A* and BFS in terms of solution plan length. The second test was intended to observe performances convergence when learning is applied in the three algorithms, i.e., when heuristics values of the visited nodes can be updated according to exploration. The results first showed that learning improves the quality of best plans obtained with the three algorithms. They moreover showed that MHSP tends to converge very quickly towards an optimal plan, while A* and BFS may stay suboptimal and unstable. Finally, the third test was designed in order to evaluate the ability of the algorithms to performs effective action selection, or partial plan selection in real-time. This test makes the decision time vary, and assesses the quality of partial plans obtained through two distances: the distance to the goal and the distance to the optimum. The results highlighted that as decision time grows up, MHSP is much faster to provide optimal plans than the two other algorithms do.

In the future, we may study several specific aspects of the presented work:
\begin{itemize}
\item First of all, we would study the possibility to perform sequences of actions instead of just one action, like algorithms such as $\gamma$-Trap or LRTS do. Indeed, instead of taking a single action between the lookahead search episodes, it applied $d$ actions to amortize the planning cost. This allows to speed up the search of a solution plan when the heuristic function is informative.
\item Moreover, since MHSP uses mean values, we also want to apply MHSP on problems in non deterministic environments and compare it to on-line MDP algorithms. 
\item Experimentations show that the first action chosen significantly impact the quality of the solution plan found in terms of lenght. For instance in blocksworld domains, choosing first a bad block to move implies to add many actions to repair this bad choice. Consequently, the idea is to allocated more time to the first reasonning step.
\item In our experiments, for each action selection stage, the tree is computed from scratch. Re-using the tree computed during the previous action selection stages is an interesting enhancement to our work. It will enable the real-time algorithms to tackle more difficult problems.
\item Our work is done in the background of real-time search, and partial plan selection. Removing the real-time constraint, and testing MHSP on problems in which full solution plans are required is an interesting research direction. However, preliminary tests show that, with almost exact heuristics, MHSP is hardly comparable to efficient and general planners using Enforced-Hill Climbing to find full solution plans on a wide range of problems.
\item Finally, learning the heuristics useful in planning by using MHSP, or another real-time algorithm, and compare them with the heuristics obtained with planning graphs is a good perspective linking the two domains of learning and planning.
\end{itemize}

\renewcommand{\refname}{References}
\newcommand{\etalchar}[1]{$^{#1}$}

\end{document}